\documentclass[letterpaper, 10 pt, conference]{ieeeconf}

\IEEEoverridecommandlockouts

\newcommand{\AVAR}{\operatorname{AV@R}}  
\newcommand{\VAR}{\operatorname{V@R}}  

\usepackage{graphicx}
\usepackage{times}
\usepackage{graphicx}
\usepackage{amsmath}
\usepackage{amssymb}
\usepackage{cite}
\usepackage{hyperref}
\usepackage{algorithmic}

\usepackage{subcaption}
\usepackage{multirow}
\usepackage{xcolor}
\usepackage{booktabs}
\usepackage{makecell}
\usepackage{mathtools}
\usepackage{bm}
\usepackage{physics}
\usepackage{subcaption}
\usepackage{url}
\usepackage[linesnumbered,ruled,vlined]{algorithm2e}
\usepackage[utf8]{inputenc}
\usepackage[normalem]{ulem}
\usepackage{xcolor}
\usepackage{tikz}
\usetikzlibrary{arrows.meta,positioning,calc,fit,shapes.geometric,shapes.multipart,shapes.symbols}
\usetikzlibrary{backgrounds}

\usepackage{caption}
\newtheorem{lemma}{Lemma}

\newtheorem{proposition}{Proposition}
\newtheorem{remark}{Remark}

\newtheorem{assumption}{Assumption}

 \newcommand{\R}{{\mathbb{R}}}

\usepackage{accents}
\newlength{\dhatheight}

\usepackage{pgf,tikz}

\usepackage[margin=1in]{geometry}

\begin{document}

\title{Conflict-Aware Active Perception and Control in 3D Gaussian Splatting Fields via Control Barrier Functions}

\author{Amirhossein Mollaei Khass, Athanasios Cosse, Vivek Pandey, Nader Motee
\thanks{
A.M. Khass, A. Cosse, V. Pandey, and N. Motee are with the Department of Mechanical Engineering and Mechanics, Lehigh University, Bethlehem, PA, 18015, USA. {\tt\small \{ammb23,asc425,vkp219,motee\}@lehigh.edu}.\endgraf
Project website: \url{https://sircesoc.github.io/Conflict_Aware_Active_Perception/}
}
}


\maketitle

\thispagestyle{empty}
\pagestyle{empty}

\maketitle

\thispagestyle{empty}
\pagestyle{empty}


\begin{abstract}
Active perception in uncertain environments requires robots to navigate safely while acquiring informative observations to reduce map uncertainty. These objectives inherently conflict, as informative viewpoints often lie near uncertain regions with higher collision risk. 
To address this challenge, we develop a conflict-aware active perception and control framework for robotic systems operating in environments represented by 3D Gaussian Splatting (3DGS). Safety is enforced using a Control Barrier Function (CBF) derived from an Average Value-at-Risk ($\AVAR$) collision-risk metric that accounts for geometric uncertainty and guarantees forward invariance of a safe set. To improve perception, we propose a risk-aware Expected Information Gain (EIG) formulation for selecting the next-best-view and introduce perception barrier functions that align the camera orientation with the local information-ascent direction. To obtain a tractable formulation for these conflicting safety and perception objectives, we propose a unified safety-critical, perception-aware quadratic program that enforces safety as a hard constraint while relaxing perception constraints through slack variables.
Simulation results demonstrate that the proposed method improves both safety and information acquisition compared to existing 3DGS-based approaches.
\end{abstract}


\vspace{-0.0cm}

\section{Introduction}
Robots operating in unknown or uncertain environments must ensure safe navigation while simultaneously improving their understanding of the surrounding scene. Achieving this objective requires actively acquiring informative observations~\cite{pandey2025efficient,jiang2024agslam,nagami2026vista,lee2024biosignal} that reduce map uncertainty while maintaining safety with respect to collision risk~\cite{khass2025active,tao2025rt,khojaste2026risk}. These objectives are inherently coupled: informative viewpoints are often located near unexplored regions where geometric uncertainty and collision risk are high. Consequently, robotic systems must balance the need for information acquisition with the requirement of safety during motion.

Control Barrier Functions (CBFs) provide a principled framework for enforcing safety constraints in control-affine systems by guaranteeing forward invariance of a safe set \cite{ames2019control,QPAmes2017,HRDCBF}. At the same time, recent advances in perception have introduced 3D Gaussian Splatting (3DGS) as an efficient point-based scene representation capable of reconstructing complex environments in real time \cite{kerbl3Dgaussians}. Due to its differentiable rendering and efficient map updates, 3DGS enables active perception strategies in which robots select informative viewpoints to maximize information gain and improve map quality \cite{Jiang2023FisherRF,jiang2024agslam,khass2025active}.

Recent works have begun exploiting 3D Gaussian Splatting (3DGS) for safe robot navigation.
Splat-Nav \cite{chen2024splatnav} models both the robot and Gaussian splats as ellipsoids and enforces safety through geometric intersection tests. 
SAFER-Splat \cite{safersplat} employs Control Barrier Functions (CBFs) to guarantee safe navigation with respect to Gaussian primitives, while \cite{tscholl2025perception} introduces a collision-cone CBF formulation for proactive collision avoidance. While these approaches provide mechanisms for safe navigation in 3DGS environments, they primarily focus on collision avoidance and do not explicitly address the interaction between safety constraints and perception objectives.

A key challenge in safety-aware active perception is the inherent trade-off between safety and perception. Informative viewpoints typically lie near unexplored or uncertain regions that may also correspond to potentially high-risk areas. Consequently, actions that maximize information gain often steer the robot toward uncertain regions, while conservative safety policies may prevent the robot from acquiring the observations required to improve the map. This fundamental conflict motivates the need for a framework that jointly reasons about perception objectives and safety-critical control.

In this paper, we propose a conflict-aware active perception and control framework for systems operating in 3D Gaussian Splatting environments. The proposed approach integrates safety and perception objectives within a unified optimization framework that explicitly accounts for map uncertainty while enabling trajectory-aware perception.

Our main contributions in this work are as follows. We develop a unified framework for safe and perception-aware navigation of robotic systems with nonlinear control-affine dynamics in environments represented by 3D Gaussian Splatting. A  safety CBF based on average value-at-risk is designed that quantifies probabilistic collision risk in the underlying 3D Gaussian field. We introduce perception CBFs that guide the system toward directions of higher expected information gain (EIG), enabling trajectory-aware active perception. Then we formulate a conflict-aware control strategy using a primal--dual CBF-QP formulation that resolves the trade-off between safety and perception by enforcing safety as a hard constraint while relaxing perception through slack variables. Extensive numerical simulations demonstrate the effectiveness of the proposed framework, improving both safety and information acquisition compared to existing 3DGS-based approaches.
\section{Problem Formulation}
\label{subsec:cbf}
This work addresses the problem of designing a control input that ensures safe navigation in an uncertain 3D Gaussian Splatting (3DGS) environment while enabling active perception to reduce map uncertainty and guide the robot toward its goal. Active perception directs the system’s sensors toward occluded and uncertain regions, while safety formulations respond to uncertainty by becoming more conservative, adding extra safety margins around these same areas. As a result, a safety–perception deadlock emerges. To address the two conflicting objectives, we propose a control method that utilizes two types of barrier functions, one for safety and one for perception. 

We consider robotic systems whose  dynamics can be modeled as  
\begin{equation}
\dot{x} = f(x) + g(x)u,
\label{eq:system}
\end{equation}
where $x \in \mathbb{R}^n$ is the state and $u \in \mathbb{R}^m$ is the control input. The vector fields $f$ and $g$ are continuous and locally Lipschitz. We define the output mappings $\pi(x) \in \mathbb{R}^d$ for the robot position and $\eta(x) \in \mathbb{S}^{d-1}$ for the unit-norm heading direction aligned with the camera viewing axis.

Given a control-affine system of the form \eqref{eq:system}, ensuring safe navigation requires maintaining the system within a set of states that are free from collisions. To formalize this notion, we define a \emph{safe set} $\mathcal{C}_s$ as the set of all states that satisfy a safety condition encoded by a continuously differentiable function $h_s$. Maintaining forward invariance of this set guarantees that the system remains safe for all time~\cite{ames2019control}. To enforce this, we employ the framework of Control Barrier Functions (CBFs), which provides a systematic way to translate state constraints into constraints on the control input. 

In addition to safety, the robot must acquire informative observations to reduce map uncertainty. 
The active perception objectives arise from two distinct geometric sources: spatial information ascent, which affects translational motion, and rotational information ascent, which influences heading. These quantities are inherently defined in different units (meters versus radians), making their integration nontrivial and often requiring careful balancing. To incorporate perception, we define two complementary objectives. First, a spatial perception barrier $h_{\pi}$ promotes alignment of the heading with the local information-ascent direction. Second, an angular perception condition $h_{\eta}$ promotes information gain through rotational motion.
However, enforcing the standard CBF inequality constraints for both safety and perception may render the problem infeasible due to their conflicting objectives. To address this issue, we introduce slack variables in the optimization formulation to relax the perception CBF constraints while maintaining the safety constraint as a hard constraint. Specifically, we consider the following optimal control strategy:
\begin{equation} 
\begin{aligned}
\underset{u, \delta_\pi, \delta_\eta}{\text{minimize}}\quad
& \frac{1}{2}\| u - u_{\mathrm{ref}} \|^2 + w_\pi \delta_\pi^q + w_\eta \delta_\eta^q \\
\text{subject to:}\quad
& \dot{h}_s \ge - \gamma_s \alpha_s(h_s) \\
& \dot{h}_{\pi} \ge - \gamma_{\pi} \alpha_{\pi}(h_{\pi}) - \delta_\pi \\
& \dot{h}_{\eta} \geq - \gamma_{\eta} \alpha_{\eta}(h_{\eta}) - \delta_\eta\\
& u_{\min} \preceq u \preceq u_{\max} \\
& \delta_\pi, \delta_\eta \ge 0
\end{aligned}
\label{eq:final_qp}
\end{equation}
in which $u_{\mathrm{ref}}$ denotes the nominal control input, with $\gamma_{\star} > 0$ are design parameters and $\dot{h}_{\star} = L_f h_{\star} + L_g h_{\star}\,u $, 
where $\star \in \{s,\pi,\eta\}$ \footnote{The common notation $L_{f}h$ is utilized for the Lie derivative of the $h(x)$ along the vector field $f(x)$.}.
The variables $\delta_\pi,\delta_\eta \ge 0$ are slack variables that relax the spatial and angular perception objectives. The parameters $w_\pi,w_\eta>0$ are the associated slack weights. The exponent $q \in \{1,2\}$ determines the relaxation profile, where $q=1$ corresponds to a sparse ($\ell_1$-norm) penalty and $q=2$ to a quadratic ($\ell_2$-norm) penalty.

In \eqref{eq:final_qp}, safety is enforced as a hard constraint, while perception objectives are relaxed through slack variables. As the safety margin decreases, the slack variables increase and perception is progressively relaxed to preserve safety.

\vspace{0.1cm}
\textbf{Problem Statement:} 
The objective is to design a safe and perception-aware optimal control policy for robot navigation in uncertain environments. Specifically, our goal is to obtain a tractable formulation of the optimization problem in \eqref{eq:final_qp} and solve it efficiently using a primal--dual CBF-QP approach.

\section{Risk-Aware Information Gain and View Selection}
\label{sec:Preliminaries}

Since our objective is perception-aware navigation, we first introduce a framework for quantifying the information gained about the environment during perception. To this end, we briefly review 3D Gaussian Splatting for environment representation and introduce a risk-aware masking mechanism that prioritizes perception in regions relevant to the robot's planned trajectory. We also introduce the risk metric used to evaluate collision risk under uncertainty. Based on this representation, we then define an information gain measure used for view selection.
We begin with a brief introduction to 3D Gaussian Splatting and the associated information gain quantification.




\subsection{Next Best View in 3D Gaussian Splatting Map}
\label{subsec:3DGS}

The environment is represented using 3D Gaussian Splatting (3DGS)~\cite{kerbl3Dgaussians}, where the scene is modeled as a finite set of Gaussian primitives $\mathcal{G} := \{g_i\}_{i=1}^{M}$, where each splat $g_i = (\mu_i, q_i, S_i, o_i, c_i, s_i) $
is parameterized by mean position $\mu_i \in \mathbb{R}^3$, quaternion $q_i \in \mathbb{R}^4$, principal scales $S_i \in \mathbb{R}_{>0}^3$, opacity $o_i \in \mathbb{R}$, base color $c_i \in \mathbb{R}^3$, and spherical harmonics $s_i \in \R^{s_i}$.

The rotation and scale define the covariance
\begin{equation}
\Sigma_i = R_i S_i S_i^\top R_i^\top.
\label{eq:splat_covariance}
\end{equation}
where $R_i$ is rotation matrix.
Since our goal is perception-aware safe navigation, we therefore select robot poses that reduce uncertainty in the 3D Gaussian Splatting (3DGS) map through active view acquisition (Next Best View, NBV).

Let $f(Y\mid(\pi,\eta);\mathcal G)$ denote the rendering model that predicts an image $Y$ from pose characterized by $(\pi,\eta)$ given Gaussian splats $\mathcal G$\footnote{We use $\mathcal G$ to denote both the set of splats and their parameters.}. 
To improve the map, we select views that maximize the expected information gain (EIG) of the 3DGS parameters~\cite{kirsch2022unifying}, approximated using Fisher information~\cite{Jiang2023FisherRF}
\begin{equation}
\mathcal I((\pi,\eta);\mathcal G)
=
\mathrm{tr}\!\left(
H''[Y\mid (\pi,\eta);\mathcal G]\,
H''[\mathcal G]^{-1}
\right),
\label{eq:eig_logdet}
\end{equation}
where $H''[Y\mid (\pi,\eta),\mathcal G]$ is the Fisher information matrix at pose given by  $(\pi,\eta)$, and $H''[\mathcal G]$ is the accumulated information from prior views. 
The NBV problem becomes
\begin{equation}
\arg\underset{(\pi,\eta)}{\text{maximize}} \; \mathcal I((\pi,\eta);\mathcal G).
\label{eq:opt_Pi_t}
\end{equation}


\subsection{Collision Risk Quantification}
\label{subsec:risk_metric}

Each Gaussian splat is treated as a local uncertain obstacle with mean $\mu_i$ and covariance $\Sigma_i$.  
Let $\pi\in\mathbb{R}^3$ denote the robot position. The signed distance to splat $g_i \sim \mathcal{N}(\mu_i, \Sigma_i)$ is
\begin{equation}
d_i(\pi)
=
\left\langle
\mu_i-\pi,
\frac{\mu_i-\pi}{\|\mu_i-\pi\|_2}
\right\rangle ,
\label{eq:def_signed_distance}
\end{equation}
where $\langle \cdot,\cdot\rangle$ denotes the Euclidean inner product, and $ \|\cdot\|_2 $ denotes the $2$-norm~\cite{liu2024riskawarefisherRF}.

\begin{lemma} \cite{liu2024riskawarefisherRF}
If the splat uncertainty is isotropic $\Sigma_i=\sigma_i^2 I_3$, 
the signed distance becomes a Gaussian random variable:
\begin{equation}
d_i(\pi) \sim \mathcal{N}(\|\mu_i - \pi\|_2,\sigma_i^2).
\label{eq:distance_random}
\end{equation}
\end{lemma}

To quantify risk of collision, we choose \emph{Average Value-at-Risk} ($\AVAR$) ~\cite{rockafellar2002cvar}.
 For a confidence level $\varepsilon \in (0,1)$, the Average Value-at-Risk ($\VAR$) of signed distance random variable $d_i^x = d_i(\pi)$ is defined as
\begin{equation}
    \mathrm{\AVAR}_{\varepsilon}(d_i^x)
    :=
    \mathbb{E}\!\left[
        d_i^x \;\middle|\; d_i^x < \mathrm{\VAR}_{\varepsilon}(d_i^x)
    \right]
    \label{eq:avar_definition}
\end{equation}
where $\mathrm{\VAR}_{\varepsilon}(d_i^x)
:=
\inf
\left\{
\zeta \in \mathbb{R}
\;\middle|\;
\mathbb{P}(d_i^x \le \zeta) \ge \varepsilon
\right\}$
is the Value-at-risk.

\subsection{Risk-Aware Next Best View Optimization}
\label{subsec:risk_aware_NBV_optimization}



In safety-critical navigation, information acquisition should prioritize regions affecting the robot's planned trajectory rather than the entire map. 
Let the planned trajectory be $\mathbf P = \{p_1,\dots,p_K\}, \ p_k\in\mathbb R^3$. For each waypoint, the $\AVAR$ defines a local risk
\[
r_{\min}(p_k)
=
\min_{g_i\in\mathcal G}
\mathrm{\AVAR}_\varepsilon\!\big(d_i(p_k)\big),
\]
which measures the safety margin between the robot and uncertain obstacles.

We define a trajectory-dependent masking radius
\begin{equation}
r_{\text{mask}}(p_k)
=
\beta_1
\exp(-\beta_2\, r_{\min}(p_k)),
\qquad
\beta_1,\beta_2>0,
\label{eq:mask_radius}
\end{equation}
so higher collision risk yields a larger observation region. 
The masked spatial region becomes
\begin{equation}
\small
\begin{aligned}
  \Pi_{\mathbf P}
=
\bigcup_{p_k\in\mathbf P}
\mathcal B(p_k,r_{\text{mask}}(p_k)),  
\end{aligned}
\label{eq:mask_region}
\end{equation}
where $\mathcal B(\cdot)$ denotes a Euclidean ball.

The set of splats inside the masked region is given by
\begin{equation}
\Pi_{\mathbf P|\mathcal G}
=
\{g_i\in\mathcal G \mid \mu_i\in\Pi_{\mathbf P}\},
\label{eq:splat_mask}
\end{equation}
with cardinality $|\Pi_{\mathbf P|\mathcal G}|=M_{\mathbf P}$.




This masking modifies the Fisher information
yielding the risk-aware EIG
\begin{equation}
\mathcal I((\pi,\eta);\Pi_{\mathbf P|\mathcal G})
=
\mathrm{tr}\!\left(
 H''[Y\mid (\pi,\eta);\Pi_{\mathbf P|\mathcal G}]
 H''_{\mathrm{prior}}(\Pi_{\mathbf P|\mathcal G})^{-1}
\right).
\label{eq:eig_fisher}
\end{equation}

The risk-aware next-best view is therefore
\begin{equation}
\arg\underset{(\pi,\eta)}{\text{maximize}} \; \mathcal I((\pi,\eta);\Pi_{\mathbf P|\mathcal G}).
\end{equation}

This formulation couples perception and safety by directing information acquisition toward uncertain regions of the map that lie near the robot's future trajectory.


\section{Control Barrier Functions Design}
\label{sec:CBF_th}

\subsection{Safety Barrier Function}
Based on the risk metric in \eqref{eq:avar_definition}, the risk of collision between the robot and a Gaussian splat $g_i \in \mathcal{G}$ is defined:
\begin{equation}
\rho_i(\pi) := \mathrm{AV@R}_{\varepsilon}\big(d_i(\pi\big)).
\end{equation}
To ensure safety, the robot must maintain a nonnegative risk margin with respect to all splats, leading to the condition  that the worst case collision risk remain nonnegative.
\begin{equation}
\min_{g_i \in \mathcal{G}} \rho_i(\pi) \ge 0.
\label{eq:minvar}
\end{equation}
This defines the safe set
\[
\mathcal{C}_s := \{x \in \mathbb{R}^n \mid \min_{g_i \in \mathcal{G}} \rho_i(\pi) \ge 0 \}.
\]
However, the minimum operator is nonsmooth and therefore not suitable for control barrier function design.
To obtain a continuously differentiable surrogate for the nonsmooth minimum operator, we introduce the following log-sum-exp approximation of the minimum collision risk.

Let $\rho_i(\pi)$, denote the risk of collision functions associated with the each Gaussian splat $g_i\in\mathcal G$. Define
\begin{equation}
h_s(\pi)
=
-\frac{1}{\beta}
\log\!\left(
\sum_{g_i\in\mathcal G}
\exp\!\big(-\beta \rho_i(\pi)\big)
\right),
\beta>0.
\label{eq:safety_barrier}
\end{equation}
Then $h_s(\pi)$ is a continuously differentiable approximation of the minimum collision risk.

\begin{proposition}
\label{le:bounded_2BF} 
For a finite set $\mathcal{G}=\{g_i\}_{i=1}^M$, the barrier \eqref{eq:safety_barrier} satisfies~\cite{sofmin}
\begin{equation}
\min_{g_i\in\mathcal G}\rho_i(\pi)-\frac{\ln (M)}{\beta}
\;\le\;
h_s(\pi)
\;\le\;
\min_{g_i\in\mathcal G}\rho_i(\pi),
\label{eq:logsumexp_bounds}
\end{equation}
\end{proposition}


\medskip
\noindent
\begin{proposition}
\label{prop:safety_barrier_properties}
Consider the safety barrier function $h_s$ as in
\ref{eq:safety_barrier}. Then the following properties hold:

\begin{enumerate}
    \item $h_s(\pi)$ is continuously differentiable function whenever each $\rho_i(\pi)$ is continuously differentiable.
    \item As $\beta\to\infty$, then $    h_s(\pi)\to \min_{g_i\in\mathcal G}\rho_i(\pi).$
    \item If $h_s(\pi)\ge0$, then $\rho_i(\pi)\ge0$ for all $g_i\in\mathcal G$, and therefore the robot is safe with respect to every Gaussian splat.
\end{enumerate}
\end{proposition}


The condition $h_s(\pi)\ge0$ implies nonnegative collision risk with respect to every splat, and therefore defines a sufficient condition for safety.

Moreover, assume the robot is subject to a bounded deceleration $a_{\max}>0$. The safety class-$\mathcal{K}$ function is defined by
\begin{equation}
\alpha_s(h_s) = \sqrt{2 a_{\max} h_s} ,\quad h_s \ge 0.
\label{eq:alpha_physical}
\end{equation}
The control input can generate a braking acceleration satisfying $\|\dot v\|\le a_{\max}.$
The choice \eqref{eq:alpha_physical} yields a physically interpretable safety constraint, linking the barrier condition directly to braking capability.

We then define the safe set as
\begin{equation}
\mathcal{C}_s := \{x \in \mathbb{R}^n \mid h_s(\pi) \ge 0 \},
\end{equation}
where $h_s(\pi)$ is a continuously differentiable safety barrier function. The system is safe if $\mathcal{C}_s$ is forward invariant~\cite{ames2019control}. To guarantee this, we impose the first-order CBF condition $\dot{h}_s \ge -\gamma_s \alpha_s(h_s),$
where $\gamma_s > 0$ is a design parameter and $\alpha_s(\cdot)$ is the class-$\mathcal{K}$ function.

\subsection{Perception Barrier Functions}
To support safe navigation in partially observed environments and reduce the risk of collision with uncertain or unseen obstacles, the robot must actively acquire informative observations to reduce map uncertainty. To this end, we construct perception barrier constraints that promote both spatial alignment with the local information-ascent direction and rotational motion toward informative viewpoints.

Let $\mathcal I(T;\Pi_{\mathbf{P}}\big|\mathcal G)$ denote the risk-aware expected information gain defined in \eqref{eq:eig_fisher}.  At the current state $x=x_0$, let
$\pi_0 = \pi(x_0)$ and $\eta_0 = \eta(x_0)$.
The local spatial information-ascent direction at $(\pi_0,\eta_0)$ is given by $\nabla_\pi \mathcal I\big|_{(\pi_0,\eta_0)}$ and its normalized direction is represented by $\mathbf d_\pi \in\mathbb R^d$ \cite{khass2025active}. 


\begin{assumption}
\label{assump:info_gradient} The information field is regular, i.e., the EIG function $\mathcal I$ is twice continuously differentiable and nonzero over the region of interest~\cite{khass2025active}.
\end{assumption}

We require the camera heading to remain aligned with the spatial information-ascent direction.
Let $\eta$ denote the camera heading direction and $\phi_{\max}\in(0,\pi/2)$ be the maximum allowable angular deviation. Define
\begin{equation}
h_{\pi}(\eta)
=
\left\langle
\eta,
\mathbf{d}_\pi\big|_{(\pi_0,\eta_0)}
\right\rangle
-
\cos(\phi_{\max}).
\label{eq:perception_barrier}
\end{equation}
Then $h_{\pi}$ defines a spatial perception barrier with superlevel set ${\mathcal{C}_\pi}:=\{x \in \mathbb{R}^n\mid h_{\pi}(\eta)\ge 0\}$ restricts the camera heading to remain within a cone of half-angle $\phi_{\max}$ around the spatial information-ascent direction $\mathbf d_\pi$ to increase expected information gain $\angle\big(\eta,\mathbf d_\pi\big)\le \phi_{\max}.$

Moreover, the perception class--$\mathcal{K}$ function is defined
\begin{equation}
\alpha_{\pi}(h_{\pi})
=
\big(1 + e^{-\tau \|\nabla_{\pi}\mathcal I\|}\big)\, h_{\pi},
\end{equation}
where $\tau > 0$ is a design parameter and $\|\nabla_{\pi}\mathcal I\|$ is evaluated at the current state.
This adaptive design accounts for perception sensitivity to expected information gain, enabling the system to selectively emphasize informative directions without compromising safety.

To account for orientation-aware information gain, we introduce an angular perception function that determines the preferred turning direction. 
Let $\mathbf d_\eta$ denote the normalized angular
information-ascent vector $\nabla_{\eta}\mathcal I$. We then define the angular perception function as
\begin{equation}
h_\eta(\eta)
:=
\left\langle
 \eta, \mathbf{d}_\eta\big|_{(\pi_0,\eta_0)}\
\right\rangle.
\label{eq:orientation_perception}
\end{equation}

We define the angular perception class–$\mathcal{K}$ function as
\[
\alpha_\eta(h_\eta)=k_\eta h_\eta, \qquad k_\eta > 0.
\]
In practice, $k_\eta$ is selected in accordance with the robot's bounded angular velocity to ensure that the prescribed recovery rate is compatible with the available turning authority.

These perception barrier objectives are incorporated into the optimization problem~\eqref{eq:final_qp} as
\begin{equation}
    \begin{aligned}
        \dot{h}_\pi & = & \left \langle   L_f \eta , \mathbf{d}_\pi  \right \rangle  \big|_{(\pi_0,\eta_0)} ~+ ~\left \langle L_g \eta , \mathbf{d}_\pi \right \rangle  \big|_{(\pi_0,\eta_0)} u \\
         \dot{h}_{\eta} & = & \left \langle  L_f \eta  , \mathbf{d}_\eta \right \rangle  \big|_{(\pi_0,\eta_0)} ~+ ~\left \langle L_g \eta, \mathbf{d}_\eta  \right \rangle  \big|_{(\pi_0,\eta_0)} u \\
    \end{aligned}
\end{equation}
 
\begin{remark}
\label{remark:adaptive_cbf}
As active perception acquires new observations, the 3DGS map $\mathcal{G}$ with parameters $\mathbf{w}$ is updated, resulting in updated information gain $\mathcal{I}(\cdot)$ and collision risk  $\rho_i(\cdot)$. Hence, the proposed barrier functions $(h_s, h_\pi, h_\eta)$ are adaptive and learned. 
\end{remark}

\subsection{KKT Perspective on Conflict-Awareness }
\label{subsec:method}

We characterize the KKT conditions \cite{boyd2004convex} to show how our proposed control algorithm in \eqref{eq:final_qp} is conflict-aware and how the weights $w_{\pi}$ and $w_{\eta}$ act as the arbitrators when safety and perception objectives collide. Let us consider the corresponding  Lagrangian function:
\vspace{-0.1cm}
\begin{equation*}
\begin{aligned} 
&\mathcal{L}(u,\delta,\lambda_s,\lambda_\pi,\nu_{\pi},\nu_\eta, \mu_i^+, \mu_i^-)
=
\frac{1}{2}\|u-u_{\mathrm{ref}}\|^2
\\
& 
\qquad+  w_{\pi} \delta_\pi^q + w_{\eta} \delta_\eta^q
\\
&\qquad
+\lambda_s\!\left(
- L_f h_s - L_g h_su - \gamma_s\alpha_s(h_s)
\right)
\\
&\qquad
+\lambda_\pi\!\left(
- L_f h_\pi - L_g h_\pi u - \gamma_{\pi} \alpha_\pi (h_\pi) - \delta_\pi
\right)
\\
&\qquad
+\lambda_\eta\!\left( 
-L_f h_{\eta} - L_g h_{\eta} u - \gamma_{\eta} \alpha_{\eta}(h_{\eta})-\delta_\eta
\right)
\\
& \qquad + (\mu^+)^\top(u - u_{\max})~ +~ (\mu^-)^\top(-u + u_{\min})
\\
&\qquad
+ \nu_{\pi}(-\delta_\pi) + \nu_{\eta} (-\delta_\eta),
\end{aligned}
\end{equation*}
where $q\in\{1,2\}$ and $\lambda_s,\lambda_\pi,\lambda_\eta,\nu_{\pi},\nu_{\eta},\mu_i^+, \mu_i^- \ge 0$ are the Lagrange multipliers associated with the safety, perception, and slack constraints, and control input constrains. For simplicity of our notations, we denote $A_{\star} = L_g h_{\star}$ and $b_{\star} = -L_f h_{\star} - \gamma_{\star} \alpha_{\star}(h_{\star})$ for each $\star \in \{s, \pi, \eta\}$. 


The stationarity conditions are characterized as
\begin{equation*}
\begin{aligned}
&u^\star = u_{\mathrm{ref}} + \lambda_s A_s^\top + \lambda_\pi A_p^\top +  \lambda_\eta A_\eta^\top - \mu^+ + \mu^-,\\
&w_1 q \delta_\pi^{q - 1} = \lambda_\pi + \nu_{\pi}, \quad \quad w_2 q \delta_\eta^{q - 1} = \lambda_\eta + \nu_{\eta}.
\end{aligned}
\end{equation*}
The first equation yields the optimal control input $u^\star$ as a correction of the nominal input $u_{\mathrm{ref}}$ along the directions $A_s^\top$, $A_\pi^\top$ and $A_\eta^\top$, which correspond to the safety and perception constraints. The complementary slackness conditions are given by
\begin{equation}
\label{eq:kkt_complementarity}
\begin{alignedat}{2}
    \lambda_s\big(A_s u^\star - b_s\big) &= 0, &\quad \mu_i^+(u_i^\star - u_{\mathrm{\max},i}) &= 0, \\
    \lambda_\pi\big(A_\pi u^\star + \delta_\pi^\star - b_\pi\big) &= 0, &\quad \mu_i^-(u_i^\star - u_{\mathrm{\min},i}) &= 0, \\
    \lambda_\eta\big(A_\eta u^\star + \delta_\eta^\star - b_\eta\big) &= 0, &\quad \nu_{\pi} \delta_\pi^\star &= 0, \\
    \nu_{\eta} \delta_\eta^\star &= 0. & &
\end{alignedat}
\end{equation}

The safety CBF requires the system to slow down, while the perception CBF pushes it to speed up. With standard hard constraints, this conflict makes the problem infeasible, i.e., no control input can satisfy both constraints. We resolve this by adding slack variables, $\delta_\pi$ and $\delta_\eta$, only to the perception constraints, while keeping safety strict. Since safety has no slack, its Lagrange multiplier $\lambda_s$ can grow unbounded, i.e., $0 \leq \lambda_s < \infty$, and dominate whenever conflicts arise. In those cases, the optimizer prioritizes safety, and the perception constraints relax through $\delta_\pi, \delta_\eta > 0$. The penalty exponent $q$ determines how strongly these violations are discouraged. 

When $q = 1$, the slack penalty is linear. The stationarity conditions give $w_{\pi} = \lambda_{\pi} + \nu_{\pi}$ and $w_{\eta} = \lambda_{\eta} + \nu_{\eta},$
with $\nu_{\pi}, \nu_{\eta} \geq 0$. This imposes upper bounds $0 \leq \lambda_\pi \leq w_{\pi}$ and $0 \leq \lambda_\eta \leq w_{\eta}$. If a violation occurs, e.g., $\delta_\pi^* > 0$, complementary slackness implies $\nu_\pi = 0$, which results in $\lambda_\pi = w_\pi$. In other words, the penalty saturates and violations are penalized aggressively at first, but after that, the penalty no longer increases, regardless of how large the violation becomes.

When $q = 2$, the slack penalty is quadratic. The KKT conditions give
$\lambda_\pi + \nu_\pi = 2w_\pi \delta^*_\pi$ and $\lambda_\eta + \nu_\eta = 2w_\eta \delta^*_\eta.$ If a conflict forces a violation, e.g., $\delta^*_\pi > 0$, complementary slackness implies $\nu_\pi = 0$, which results in 
$\lambda_\pi = 2w_\pi \delta^*_\pi.$
In this case, the multiplier scales with the size of the violation. Small violations produce a weak response, while larger violations lead to proportionally stronger pushback. As a result, the perception constraint relaxes smoothly under mild conflicts but increasingly resists as the violation grows, effectively acting as a proportional feedback on the constraint violation.

\section{Experiment}
The proposed framework applies to a broad class of control-affine robotic systems through the output mappings $\pi(x)$ and $\eta(x)$. In this work, we study two model dynamics systems and validate our proposed method and compare it with recent baselines.

\begin{remark}[Physical Extent of the Robot]
The above formulation treats the robot as a point mass. To account for its physical geometry, we consider an inflated collision volume $V_r \subset \mathbb{R}^3$ with radius $r_v$ contains all potential rotations of robot. The Equation~\eqref{eq:avar_definition} is then evaluated with respect to this inflated volume.~\cite{liu2024riskawarefisherRF}
\end{remark}

\subsection{Case Study I: Double integrator safety barrier}
\label{subsec:case1}
In this case study, we validate the proposed safety barrier function on a double-integrator model and compare it against existing 3DGS-based safety baseline. The SAFER-Splat~\cite{safersplat} constructs a CBF from the minimum Euclidean distance between the robot and each Gaussian ellipsoid and then applies a second-order CBF constraint. 
To ensure a fair comparison, we use the same 3DGS maps trained with Splatfacto~\cite{Xu2024SplatfactoWAN} in four scenes including Stonehenge ($100\text{K}$ Gaussian splats), Statues ($200\text{K}$), Flightgate ($300\text{K}$), and Adirondacks ($500\text{K}$), with the same PD nominal controller and model dynamics used in CBF which is given by~\cite{droneCBF,safersplat}
\begin{equation}
\dot{x}
=
\begin{bmatrix}
\dot p\\
\dot {\textbf{v}}
\end{bmatrix}
=
\begin{bmatrix}
0&I\\
0&0
\end{bmatrix}
\begin{bmatrix}
p\\
 {\textbf{v}}
\end{bmatrix}
+
\begin{bmatrix}
0\\
I
\end{bmatrix}u,
\label{eq:case1_doubleint}
\end{equation}
where $p\in\mathbb{R}^3$ is the position, $ {\textbf{v}}\in\mathbb{R}^3$ is the velocity, and $u\in\mathbb{R}^3$ is the commanded acceleration.
The proposed safety barrier $h_s$~\eqref{eq:safety_barrier} depends only on the robot position. Since the control input acts on acceleration, the barrier has relative degree two.
\begin{table}[t]
\centering
\scriptsize
\setlength{\tabcolsep}{5pt}
\renewcommand{\arraystretch}{1.02}
\begin{tabular}{llcccc}
\toprule
\textbf{Method} & \textbf{Scene} & \shortstack{\textbf{Safe}\\\textbf{Rate}} & \shortstack{\textbf{Comp.}\\\textbf{Time} $\downarrow$} & \shortstack{\textbf{Min.}\\\textbf{Dist.} $\uparrow$} & \shortstack{\textbf{Control}\\\textbf{Dev.}} \\
\midrule

\multirow{4}{*}{\shortstack[l]{SAFER-Splat}}
& Stonehenge  & 100\% & 30.9 ms & 0.086 & 0.069 \\
& Statues    & 99\%  & 45.0 ms & 0.141 & 0.028 \\
& Flightgate     & 99\%  & 60.3 ms & 0.086 & 0.062 \\
& Adirondacks  & 100\% & 186 ms  & 0.079 & 0.046 \\
\midrule

\multirow{4}{*}{\shortstack[l]{$\AVAR$-3D\\$\epsilon=0.90$}}
& Stonehenge & 100\% & 2.64 ms & 0.120 & 0.105 \\
& Statues    & 99\%  & 2.68 ms & 0.169 & 0.104 \\
& Flightgate     & 98\%  & 2.72 ms & 0.150 & 0.098 \\
& Adirondacks  & 98\%  & 2.79 ms & 0.111 & 0.098 \\
\midrule

\multirow{4}{*}{\shortstack[l]{$\AVAR$-3D\\$\epsilon=0.75$}}
& Stonehenge & 100\% & 1.63 ms & 0.106 & 0.070 \\
& Statues    & 99\%  & 1.49 ms & 0.158 & 0.024 \\
& Flightgate     & 99\%  & 1.50 ms & 0.143 & 0.064 \\
& Adirondacks  & 99\%  & 1.84 ms & 0.075 & 0.044 \\
\bottomrule
\end{tabular}
\caption{Performance comparison across 3DGS scenes. The proposed $\AVAR$-based method achieves comparable safety with significantly reduced computation time and improved minimum safety margins.}
\label{tab:case1_results}
\end{table}

The resulting safety HOCBF~\cite{HRDCBF} constraint derives as: 
\begin{equation}
\small
\begin{aligned}
\nabla_p h_s(p)^\top u
&\ge -{\mathbf v}^\top \nabla_p^2 h_s(p)\,{\mathbf v}  \\
&\quad -\gamma_s\alpha_s\!\left(\nabla_p h_s(p)^\top {\mathbf v}+\alpha_s\!\left(h_s(p)\right)\right).
\label{eq:case1_explicit_hocbf}
\end{aligned}
\end{equation}
where
\[
\begin{aligned}
\nabla_p h_s(p)=\sum_{i=1}^M\frac{e^{-\beta \rho_i(p)}}{\sum_{j=1}^M e^{-\beta \rho_j(p)}}\,\nabla_p \rho_i(p),\qquad
\end{aligned}
\]
and
\[
\small
\begin{aligned}
&\nabla_p^2 h_s(p)=\sum_{i=1}^M \frac{e^{-\beta \rho_i(p)}}{\sum_{j=1}^M e^{-\beta \rho_j(p)}}\,\nabla_p^2 \rho_i(p) 
\\
&-\beta\sum_{i=1}^M\frac{e^{-\beta \rho_i(p)}}{\sum_{j=1}^M e^{-\beta \rho_j(p)}}\big(\nabla_p\rho_i(p)-\nabla_p h_s(p)\big)\big(\nabla_p\rho_i(p)\big)^\top.
\end{aligned}
\]

The optimal control input is then obtained from
\begin{equation}
u^\star=\arg\underset{u}{ \textrm{minimize}} \ \frac{1}{2}\|u-u_{\mathrm{ref}}\|_2^2
\quad
\text{s.t.}\ \eqref{eq:case1_explicit_hocbf}.
\label{eq:case1_qp}
\end{equation}

\begin{figure}[t]
    \centering
\includegraphics[width=0.38\textwidth,height=0.17\textheight,
    trim={11cm 5.5cm 10.1cm 3.5cm},clip]{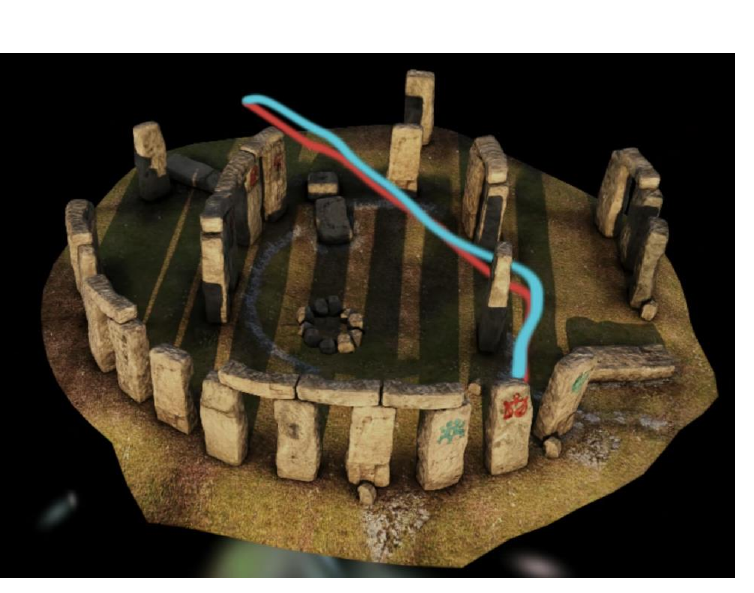}
    \caption{Comparison of navigation trajectories in Stonehenge environment between the proposed method (blue), which maintains a larger safety margin, and SAFER-Splat (red).}
    \label{fig:case1_pathes}
\end{figure}

Table~\ref{tab:case1_results} indicates the results from $100$ trajectory simulations. The proposed $\AVAR$-based safety barriers with $\epsilon=0.75,\ 0.90$ achieves comparable safety to SAFER-Splat while significantly reducing computation time, yielding $\sim (30\times)- (50\times)$ speedup across all scenes.
The parameter $\epsilon$~\eqref{eq:avar_definition} indicates the safety confidence level metric. A higher confidence level ($\epsilon=0.90$) results in a more conservative risk quantification, effectively enlarging the perceived obstacle regions induced by the Gaussian splats. In cluttered environments with dense 3DGS representations, this conservativeness can restrict the set of admissible motions, particularly in narrow passages or when the robot is initialized close to obstacles, leading to a slight reduction in success rate. 
In contrast, a moderate setting ($\epsilon=0.75$) relaxes this conservativeness and preserves feasible navigation behaviors while maintaining safety. Importantly, both settings consistently achieve larger minimum distances than SAFER-Splat. This behavior follows directly from the $\AVAR$ formulation. A qualitative comparison of the resulting trajectories is shown in Figure~\ref{fig:case1_pathes}, highlighting the improved safety margin.
Control deviation is not a reliable indicator of safety, as all approaches rely on the same nominal PD controller for goal tracking.
A recent approach~\cite{tscholl2025perception} defines a barrier directly on $(p,v)$ reports a $\sim\!3\times$ reduction in planning time compared to SAFER-Splat. However, since the implementation is not publicly available, a direct comparison is not possible.
\subsection{Case Study II: Unicycle Dynamics}
\label{subsec:case2}
We now evaluate the proposed framework, including safety and spatial perception barrier functions, on a ground robot with unicycle dynamics. This case study highlights the interaction between safety and perception, and demonstrates the proposed conflict-aware control mechanism.
The ground robot with unicycle dynamics is modeled as
\begin{equation}
\dot{x}
=
\begin{bmatrix}
\dot p\\
\dot\theta
\end{bmatrix},
\qquad\dot p=
\begin{bmatrix}
{\textbf{v}}\cos\theta\\
{\textbf{v}}\sin\theta
\end{bmatrix},
\qquad
\dot\theta=\omega,
\label{eq:case2_unicycle}
\end{equation}
where $x=(p,\theta)\in SE(2)$, with $\pi(x) = p\in\mathbb{R}^2$ denoting the robot position and $\theta\in\mathbb{S}^1$ the heading angle, and $u=( {\textbf{v}},\omega)^\top\in\mathbb{R}^2$ denotes the linear and angular velocities.
The viewing direction is aligned with the robot heading and is given by
$\eta:= \eta(x) =[\cos\theta,\ \sin\theta]^\top$.
The robot operates on a planar manifold with fixed sensing height and the environment is represented by a 3D Gaussian Splatting (3DGS) map. Under these dynamics, the safety barrier~\eqref{eq:safety_barrier} and the perception barrier \eqref{eq:perception_barrier} have relative degree one.

To make the perception barrier more explicit, let the normalized information-ascent direction be $\mathbf{d}_\pi|_{(\pi_0,\eta_0)}:= n=[n_1,n_2]^\top$ and $\|n\|=1$.

The perception barrier \eqref{eq:perception_barrier} can be written as
\begin{equation}
\begin{aligned}
h_\pi
= n_1 \cos\theta + n_2 \sin\theta-\cos\phi_{\max}.
\label{eq:hp_explicit}
\end{aligned}
\end{equation}

\begin{figure}[t]
    \centering
\includegraphics[width=\linewidth,height=0.24\textheight, trim={5.5cm 3.25cm 7.8cm 0.9cm},
    clip]{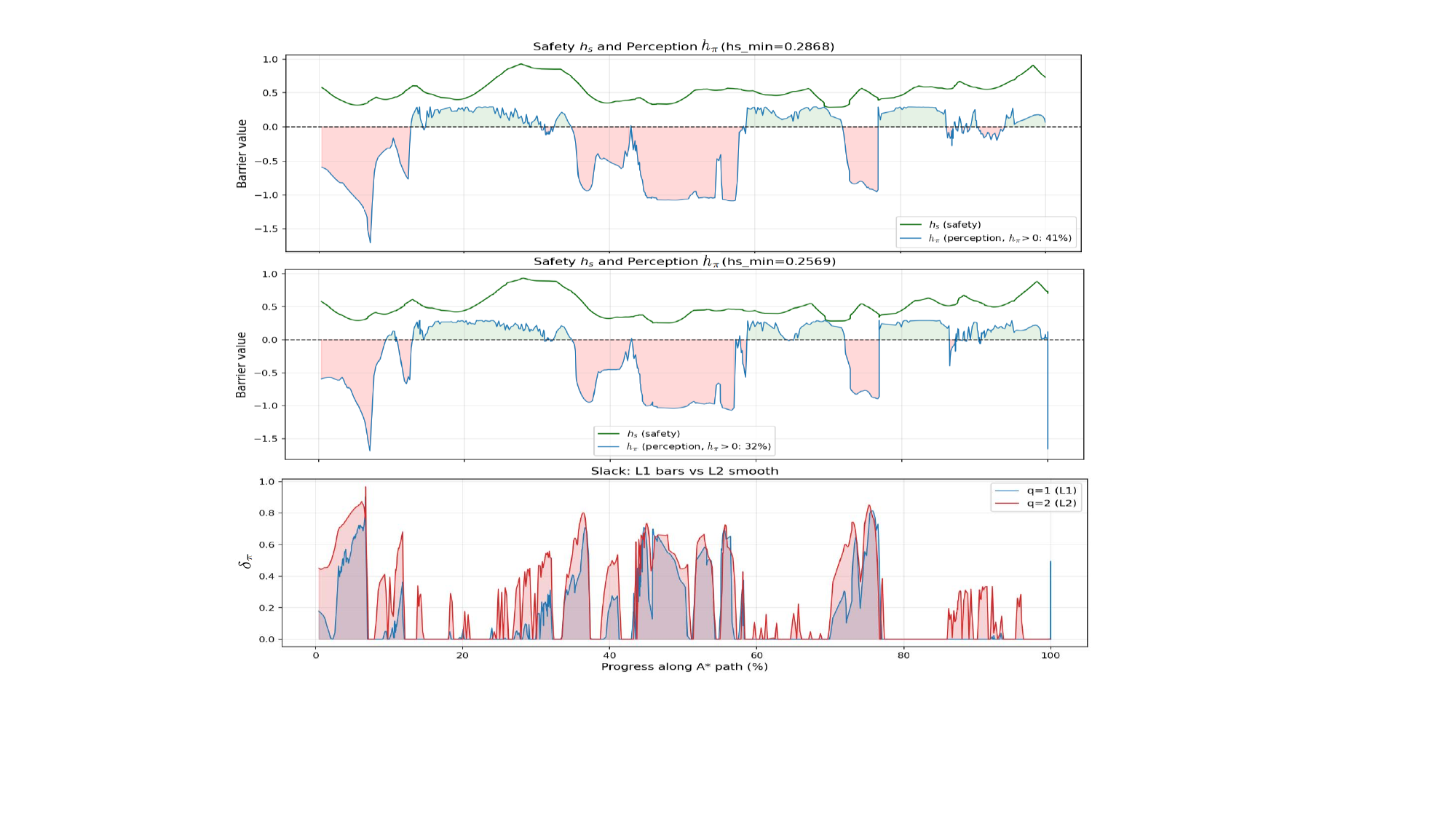}
    \caption{conflict-aware safety-perception behavior in Case Study II. The top and middle plots show the safety and perception barriers for $q=2$ and $q=1$, respectively. When $h_s$ decreases toward zero, the slack variable activates and relaxes the perception constraint. The bottom plot shows the corresponding slack $\delta_\pi$ comparing the effect of $q=1$ and $q=2$.}
    \label{fig:plotslack}
    \vspace{-0.5 cm}
\end{figure}

\begin{lemma}[Unicycle dynamics Perception barrier]
\label{le:uniperception}
 For the unicycle kinematics \eqref{eq:case2_unicycle}, the  derivative of the perception barrier function given by
\begin{equation}
\dot h_\pi =  \omega\,(J\eta)^\top n
\ 
\label{eq:hp_dot_geom}
\end{equation}
where $J\eta=\begin{bmatrix}-\sin\theta, \cos\theta\end{bmatrix}$
Equivalently, the rotational term becomes
\begin{equation}
\begin{aligned}
(J\eta)^\top n
=
n_2\cos\theta - n_1\sin\theta.
\label{eq:side_indicator}
\end{aligned}
\end{equation}   
\end{lemma}

Hence, \eqref{eq:side_indicator} acts as a signed side indicator. $(J\eta)^\top n>0$ means that the information direction lies to the \emph{left} of the current heading, whereas $(J\eta)^\top n<0$ means that it lies to the \emph{right}. Therefore, the sign of the coefficient multiplying $\omega$ directly tells the controller which turning direction increases the barrier value.
Moreover, For the unicycle dynamics, the safety barrier functions $h_s$ have relative degree one, and hence the corresponding CBF constraint $\dot{h_s}$ is calculated through their first-order derivatives.
The optimal control input is obtained by solving the conflict-aware CBF quadratic program in Equation~\ref{eq:final_qp} with the safety CBF constraint and spatial perception CBF constraint.

We evaluate the proposed conflict-aware framework in an uncertain environment where the robot incrementally updates its 3DGS map from online observations. At each time step, the robot recomputes both the safety and perception barrier functions using the current map belief and solves the CBF-QP to generate control inputs.

\begin{figure}[t] 
    \centering
\includegraphics[width=\linewidth,height=0.13\textheight, trim={4.5cm 6.25cm 6.2cm 4.5cm},
    clip]{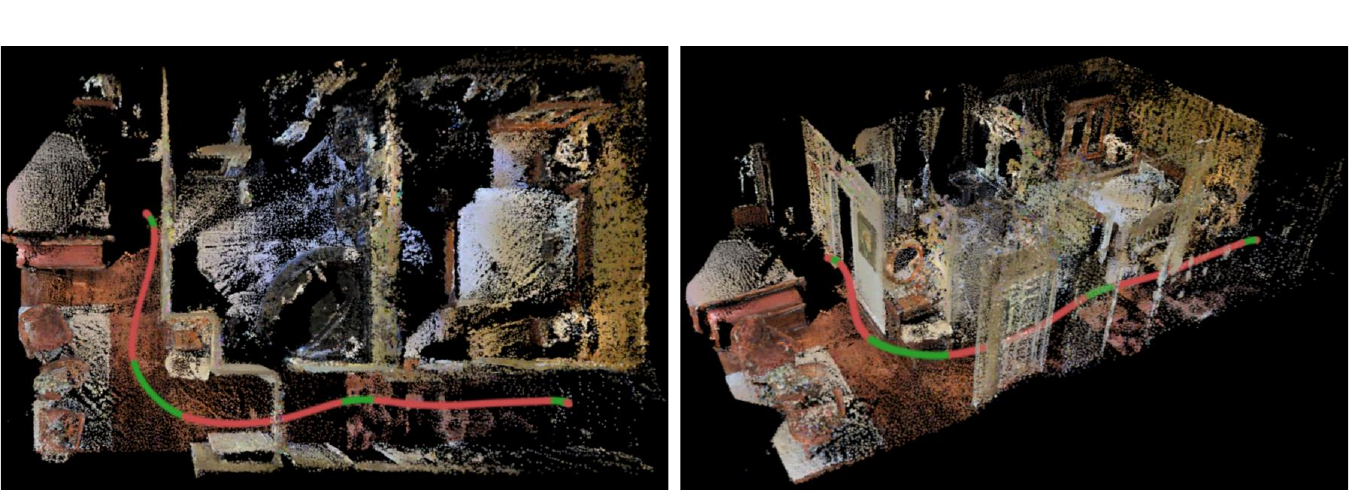}
    \caption{Trajectory for Case Study II ($q=2$) where the robot reduces map uncertainty while navigating. Green segments indicate active perception ($h_\pi>0$) with no conflict. Red segments indicate perception relaxation ($h_\pi\leq 0$), either due to safety constraints or low information gain in the trajectory relevant region.
    }
    \label{fig:slacktrack}
    \vspace{-0.5 cm}
\end{figure}

The results in Figure~\ref{fig:plotslack} illustrate the conflict-aware active perception and safety along the trajectory in Figure~\ref{fig:slacktrack}. In safe regions, $h_s$ remains well above zero, the perception constraint is active and the robot aligns its motion with the information-ascent direction, resulting in positive values of $h_\pi$ and increased information acquisition. As the robot approaches cluttered or narrow regions, the safety barrier $h_s$ decreases where the optimization prioritizes safety by activating the slack variable $\delta_\pi$ and relaxes the perception constraint to ensure collision avoidance.

The choice of $q$ changes how this relaxation occurs. For $q=1$, the penalty produces sparse activations of $\delta_p$, whereas $q=2$ yields smoother relaxation over longer portions of the trajectory. In our experiments, $q=2$ leads to slightly larger minimum safety margins, consistent with a more gradual reduction of perception influence before safety becomes critical. The slack weight also plays a key role as increasing it enforces perception more strongly and drives the robot toward uncertain regions, but if it is overly large, the robot may not find a dynamically feasible control input to recover safe motion near obstacles.

\subsection{Case Study III: Unicycle Dynamics angular Perception constraint}
\label{subsec:case3}
We extend Case Study II by adding the angular perception constraint. Under unicycle system~\eqref{eq:case2_unicycle}, angular perception barrier constraint become
\begin{equation}
  \dot{h}_\eta =  \langle
 \eta^\perp, \mathbf{d}_\eta\big|_{(\pi_0,\eta_0)}\
\rangle\omega
\end{equation}
where $\eta =[\cos\theta,\ \sin\theta]^\top$

\begin{figure}[t]
    \centering
\includegraphics[width=\linewidth,height=0.22\textheight, trim={6.9cm 5.2cm 6.5cm 4.0cm},
    clip]{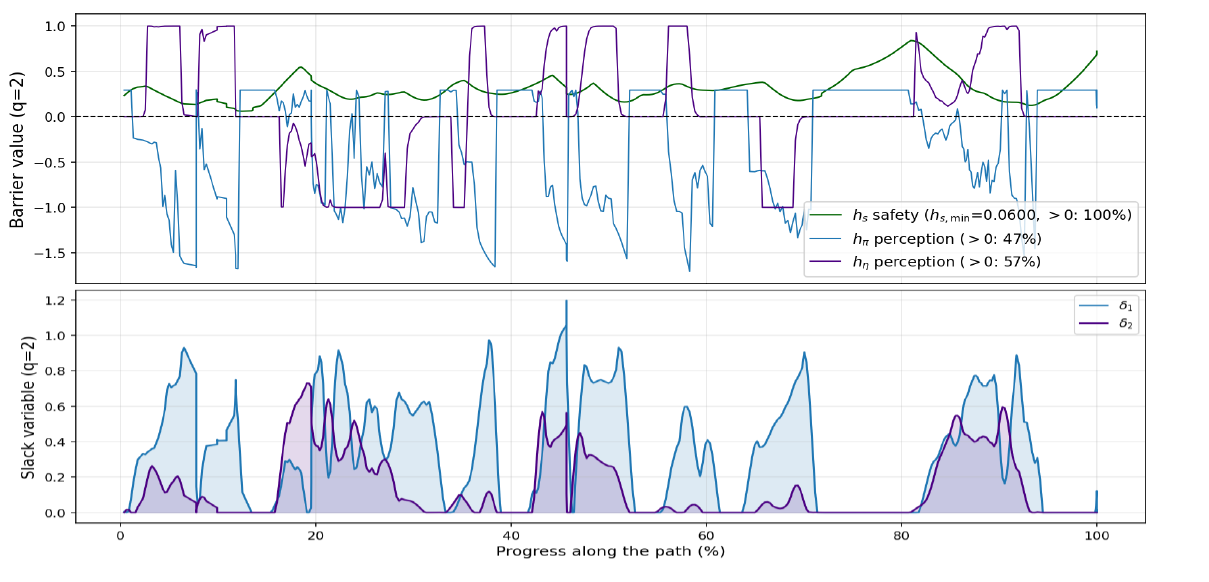}
    \caption{Conflict behaviors in Case Study III. The top plot shows the evolution of $h_s$, $h_\pi$, and $h_\eta$. When conflicts happen corresponding perception constraints are relaxes. The bottom plot shows the slack variables $\delta_\pi$ and $\delta_\eta$, indicating adaptive relaxation of spatial and angular perception to preserve safety.}
    \label{fig:plotslacktwo}
\end{figure}

\begin{figure}[t]
    \centering
\includegraphics[width=\linewidth,height=0.21\textheight, trim={1.6cm 16.0cm 2.3cm 0.7cm},
    clip]{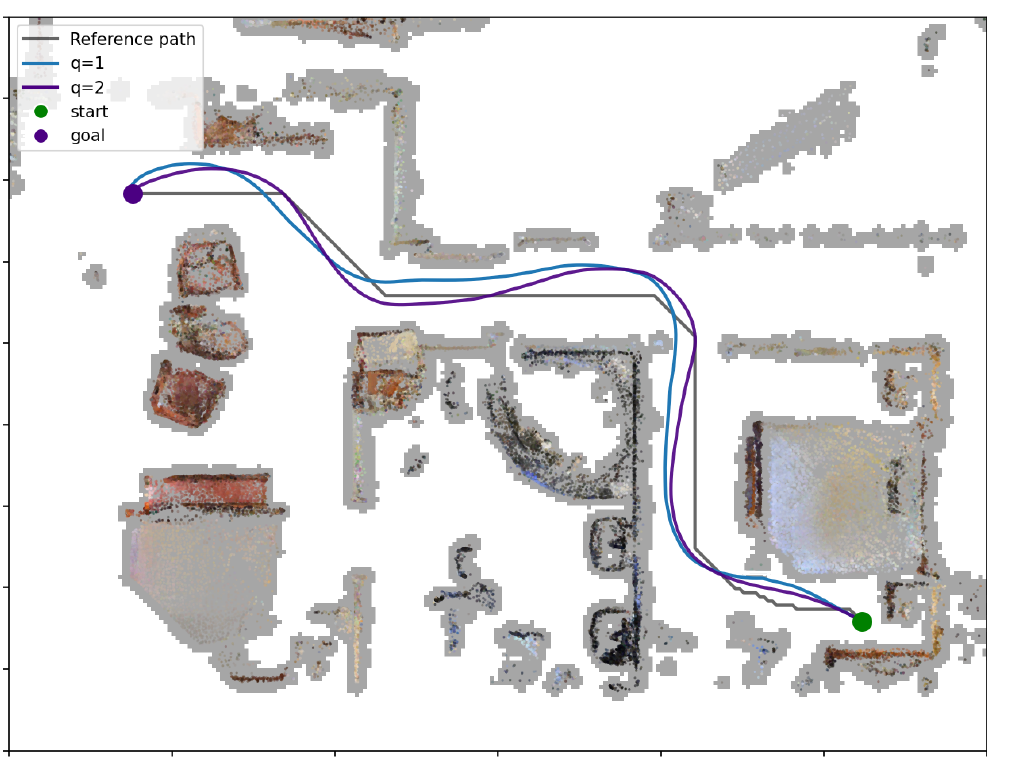}
    \caption{Trajectories for Case Study III correspond to Figure~\ref{fig:plotslacktwo}. }
    \label{fig:slackpathtwo}
    \vspace{-0.5 cm}
\end{figure}

The resulting control input is obtained from~\eqref{eq:final_qp} yielding a coupled perception framework with both spatial and angular objectives.
 As shown in Figure~\ref{fig:plotslacktwo}, safety is preserved while the slack variables adaptively relax one or both perception constraints when conflicts arise, including conflicts between the two perception barrier objectives. Figure~\ref{fig:slackpathtwo} shows that adding angular perception improves viewpoint selection by explicitly steering the heading toward informative orientations.

\section{Conclusion}
In this paper, we presented a conflict-aware active perception and control framework for robots operating in environments represented by 3D Gaussian Splatting. The proposed approach integrates safety and perception objectives within a unified optimization framework by combining a risk-aware CBF based on an $\AVAR$ collision metric with a perception barrier that promotes informative viewpoint selection. This formulation enables robots to safely navigate uncertain environments while actively improving map quality through trajectory-aware perception. Simulation results demonstrate that the proposed method improves both safety and information acquisition compared to existing 3DGS-based navigation approaches while maintaining comparable estimation accuracy. Future work will focus on extending the framework to systems with relative degree greater than or equal to two using exponential CBFs, implementing the approach on real robotic platforms, and extending the formulation to dynamic environments and multi-robot active perception.
\begin{figure}[t]
    \centering
\includegraphics[width=\linewidth,height=0.22\textheight, trim={7.0cm 4.25cm 9.0cm 2.5cm},
    clip]{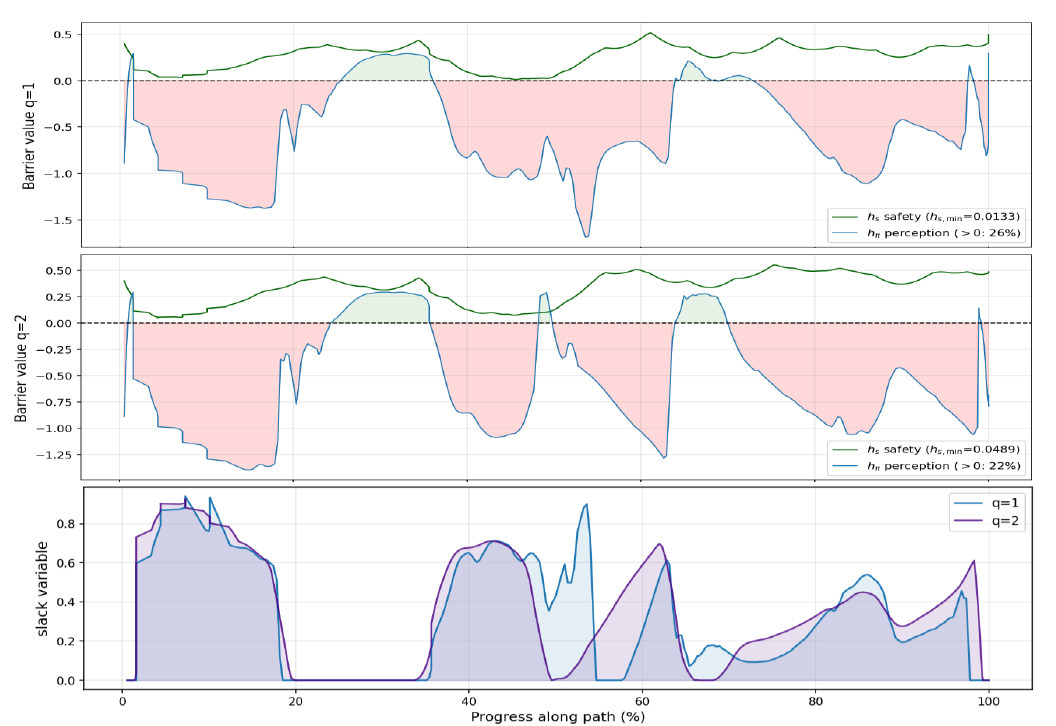}
    \caption{ Results follow the same simulation as Figure~\ref{fig:plotslack} as Case Study II with higher slack weight to penalize the perception in the conflict}
    \label{fig:plotslack2}
    \vspace{-0.5 cm}
\end{figure}

{\footnotesize
\bibliographystyle{IEEEtran}
\bibliography{references}
}

\end{document}